\documentclass[10pt,twocolumn,letterpaper]{article}
\RequirePackage[table]{xcolor} % 放在文档最前面
\usepackage{makecell}
\usepackage{cvpr}              % To produce the CAMERA-READY version
\usepackage{times}
\usepackage{epsfig}
\usepackage{graphicx}
\usepackage{caption}
\usepackage{amsmath}
\usepackage{amssymb}
\usepackage{booktabs}
\usepackage{multirow}
\usepackage{enumitem}
\usepackage{diagbox} % 引入diagbox宏包用于绘制斜线单元格
\usepackage{array} % 引入array宏包以便控制表格列宽
\definecolor{cvprblue}{rgb}{0.21,0.49,0.74}
\usepackage[pagebackref,breaklinks,colorlinks,allcolors=cvprblue]{hyperref}

\def\paperID{4543} % *** Enter the Paper ID here
\def\confName{CVPR}
\def\confYear{2025}

\title{S2Gaussian: Sparse-View Super-Resolution 3D Gaussian Splatting}

\author{
		Yecong Wan$^{1}$, 
		Mingwen Shao$^{1}$\footnotemark[1] , 
		Yuanshuo Cheng$^{1}$, 
		Wangmeng Zuo$^{2}$ \\
		$^{1}$Qingdao Institute of Software, College of Computer Science and Technology, \\China University of Petroleum (East China) \\
		$^{2}$Faculty of Computing, Harbin Institute of Technology \\
		smw278@126.com,
		\{yecongwan, cys1294414023, cswmzuo\}@gmail.com \\
}
\begin{document}
\twocolumn[{% 
	\renewcommand\twocolumn[1][]{#1}% 
	\maketitle 
	\begin{center}
		\centering 
		\includegraphics[width=\textwidth]{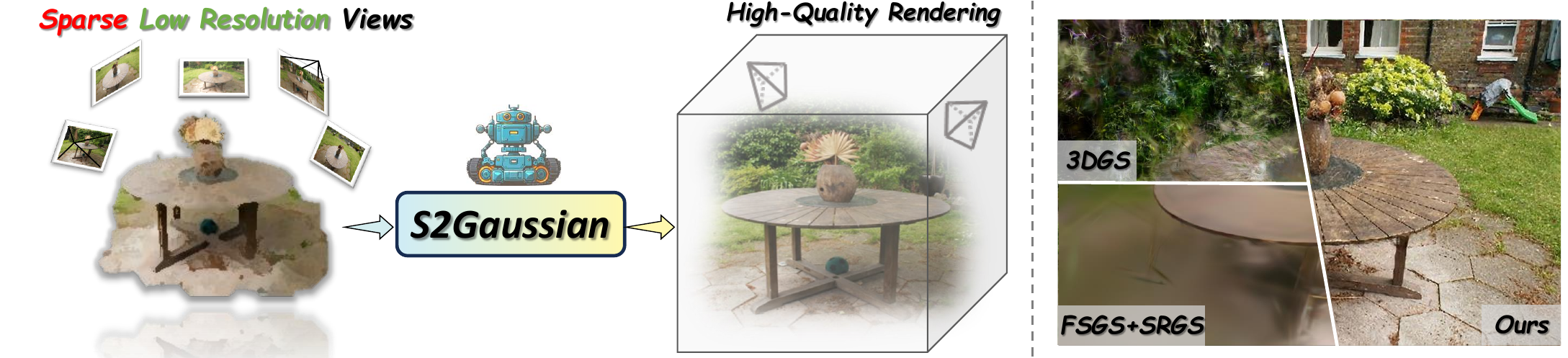}
		\captionof{figure}{We present \textbf{S2Gaussian}, a novel framework capable of reconstructing high-quality 3D scenes for immersive rendering with only sparse and low-resolution input views.
			S2Gaussian demonstrates superior performance and yields high-fidelity and high-resolution scene reconstruction with sharp geometric and detailed textures, thus enjoying better functionality and practicality in realistic applications.}
	\end{center}% 
}]

\maketitle
\footnotetext[1]{$^{*}$ Corresponding authors.}
\begin{abstract}
In this paper, we aim ambitiously for a realistic yet challenging problem, namely, how to reconstruct high-quality 3D scenes from sparse low-resolution views that simultaneously suffer from deficient perspectives and clarity. Whereas existing methods only deal with either sparse views or low-resolution observations, they fail to handle such hybrid and complicated scenarios. To this end, we propose a novel Sparse-view Super-resolution 3D Gaussian Splatting framework, dubbed S2Gaussian, that can reconstruct structure-accurate and detail-faithful 3D scenes with only sparse and low-resolution views. The S2Gaussian operates in a two-stage fashion. In the first stage, we initially optimize a low-resolution Gaussian representation with depth regularization and densify it to initialize the high-resolution Gaussians through a tailored Gaussian Shuffle Split operation. In the second stage, we refine the high-resolution Gaussians with the super-resolved images generated from both original sparse views and pseudo-views rendered by the low-resolution Gaussians. In which a customized blur-free inconsistency modeling scheme and a 3D robust optimization strategy are elaborately designed to mitigate multi-view inconsistency and eliminate erroneous updates caused by imperfect supervision. Extensive experiments demonstrate superior results and in particular establishing new state-of-the-art performances with more consistent geometry and finer details. Project Page \url{https://jeasco.github.io/S2Gaussian/}.
\end{abstract}

%%%%%%%%% BODY TEXT}
\section{Introduction}
The rapid advancement of virtual reality and metaverse technologies has significantly heightened the demand for realistic 3D scene reconstruction, which holds vast potential across various application domains, including medicine, education, entertainment, etc. 
%Over the past few decades, substantial research and development efforts \cite{mildenhall2021nerf,muller2022instant,kerbl20233d,yu2024mip} have been devoted to tackling this challenge, aiming to produce practical solutions that enable more effective novel view rendering for immersive experiences.
Among the recent notable developments, Neural Radiance Fields (NeRF) \cite{mildenhall2021nerf} have demonstrated remarkable capability in synthesizing photorealistic images; however, they still face considerable rendering and training costs, despite ongoing improvements \cite{chen2022tensorf, fridovich2022plenoxels, garbin2021fastnerf, muller2022instant}. 
More recently, 3D Gaussian Splatting
(3DGS) \cite{kerbl20233d} has emerged as a prominent approach, characterized by its high quality, rapid reconstruction speed, and support for real-time rendering. Subsequent research efforts \cite{yu2024mip, cheng2024gaussianpro, wu20244d, lu2024scaffold, tang2023dreamgaussian,chen2024survey} have focused on extending the applicability of 3DGS across various scenarios. 

Unfortunately, these methods typically rely heavily on well-captured dense and high-resolution images for impressive novel view synthesis, which is cumbersome and sometimes impractical in real-world applications. To address this challenge, a variety of approaches have emerged that focus on reconstructing 3D scenes from sparse \cite{deng2022depth, niemeyer2022regnerf, yang2023freenerf, wang2023sparsenerf, chung2024depth, zhu2025fsgs, li2024dngaussian, xiong2023sparsegs} and low-resolution views \cite{han2023super, lee2024disr, yoon2023cross, feng2024srgs, yu2024gaussiansr, shen2024supergaussian}, yielding worth-celebrating successes. However,
these two obstacles, i.e., sparsity and low resolution, have long been treated as isolated problems and tackled separately, whereas input views in practical applications often suffer from both deficient perspectives and clarity. Particularly in contexts such as robotics and internet-collected imagery, where the available views are usually sparse and compounded by low resolutions due to environmental constraints and hardware transfer limitations. 
This interplay of sparsity and low-resolution poses significant challenges for existing reconstruction frameworks. 
Furthermore, merely incorporating existing methods remains subject to inherent incompatibility, as super-resolution necessitates dense supervision to recover fine details while sparse-view regularization not only fails to introduce details but tends to lead smoothing.

To tackle the aforementioned problems, in this paper, we propose a novel 3D reconstruction framework S2Gaussian, capable of reconstructing
geometrically precise and richly detailed 3D scenes with only sparse and low-resolution views. Specifically, S2Gaussian consists of two main phases: the high-resolution Gaussians (HR GS) initialization stage and the HR GS optimization stage. In the HR GS initialization stage, we first optimize a low-resolution Gaussian representation using the sparse views as well as the estimated depth information. The optimized low-resolution Gaussians are then utilized to initialize the high-resolution Gaussians through a tailored Gaussian Shuffle Split operation which can provide more compact Gaussian primitives to facilitate fine-grained details reconstruction in high-resolution scenes. In the HR GS optimization stage, we use super-resolved images generated
by pre-trained super-resolution model to refine the high-resolution Gaussians, leveraging both the original sparse views and the pseudo-views rendered from the low-resolution Gaussians. Notably, a customized blur-free inconsistency modeling scheme and a 3D robust optimization strategy are introduced to mitigate the impacts of multi-view inconsistency and imperfect supervision, ultimately leading to more detailed and higher-quality scene reconstruction.
Comprehensive experiments substantiate 
the superiority of our S2Gaussian beyond alternatives on a variety of benchmarks.

In conclusion, the main contributions are as follows:
\begin{itemize}[noitemsep]
	\item We propose an innovative two-stage framework, termed S2Gaussian, designed to reconstruct 3D scenes that are both structurally accurate and rich in detail using only sparse and low-resolution input views.
	\item A dedicated Gaussian Shuffle Split operation is designed to initialize more compact Gaussian primitives for representing fine-grained details and textures in high-resolution scenes.	
	\item We introduce a tailored blur-free inconsistency modeling scheme alongside a 3D robust optimization strategy to address multi-view inconsistencies and rectify erroneous updates arising from imperfect supervision.
	\item Our S2Gaussian reconstructs more fine-grained and high-quality Gaussian representations significantly outperforming existing methods across multiple benchmarks.
\end{itemize}
\vspace{1pt}

\section{Related Work}
\subsection{Novel View Synthesis using Radiance Fields}
\vspace{-1pt}
Techniques for novel view synthesis typically involve learning a 3D representation using a limited number of input views and generating images from arbitrary novel perspectives. Recent advancements in Neural Radiance
Fields (NeRF) \cite{mildenhall2021nerf} has demonstrated encouraging progress in this field, which learns an implicit neural scene representation for novel view synthesis via coordinate-based neural networks and volume rendering function. Subsequent research has focused extensively on enhancing NeRF's rendering quality \cite{barron2021mip,barron2022mip,barron2023zip,verbin2022ref}, improving efficiency \cite{chen2022tensorf,fridovich2022plenoxels,garbin2021fastnerf,muller2022instant}, advancing scene understanding \cite{kerr2023lerf,zheng2024surface,zhi2021place}, and facilitating 3D content generation \cite{hollein2023text2room,poole2022dreamfusion,seo2023let}. Unfortunately, while NeRF significantly improves the quality of novel view rendering, its expensive training time as well as slow rendering speed hinder broader practical applications. Recently, Kerbl et al. \cite{kerbl20233d} introduced a groundbreaking approach with their 3D Gaussian Splatting (3DGS) method, boosting rendering efficiency by utilizing explicit Gaussian representations combined with differentiable rasterization techniques. 
Building upon 3D Gaussian Splatting representation, tremendous following efforts \cite{yu2024mip,cheng2024gaussianpro,wu20244d,lu2024scaffold,tang2023dreamgaussian} concentrate extending 3DGS under various scenarios. 
Nevertheless, these methods typically require dense and high-quality input views for impressive novel view synthesis, and they will tend to overfit the training data and compromise representations when provided with only sparse or degraded views, leading to a substantial decline in performance for the intended scene reconstruction. 
\vspace{-2pt}
\subsection{Sparse Novel View Synthesis}
\vspace{-2pt}
To enhance the availability in practical usage, numerous algorithms have been developed to surmount the reliance on dense perspectives. For instance,
Depth-NeRF \cite{deng2022depth} and RegNeRF \cite{niemeyer2022regnerf} advocate utilizing depth constraints to improve the accuracy of the sparse scene representation. FreeNeRF \cite{yang2023freenerf}, on the other hand, employs a frequency regularization strategy to manage the frequency range and penalize density fields near the camera.
Additionally, SparseNeRF \cite{wang2023sparsenerf} leverages depth priors from inaccurate observations, implementing local depth ranking and spatial continuity constraints to bolster neural radiance fields.
More recently, a variety of 3DGS-based approaches have emerged \cite{chung2024depth,zhu2025fsgs,li2024dngaussian,xiong2023sparsegs}, paving the way for new blueprints in the field. These methods typically focus on various strategies that utilize depth prior to constrain the optimization of Gaussian primitives, such as SfM depth alignment \cite{chung2024depth}, Pearson correlation
depth distribution loss \cite{zhu2025fsgs}, global-local depth normalization \cite{li2024dngaussian}, and patch-based depth correlation \cite{xiong2023sparsegs}.
Unlike these approaches, we spotlight more arduous sparse reconstruction under low-resolution conditions, where available information is more scarce.

\begin{figure*}[t]
	\begin{center}
		\includegraphics[width=\linewidth]{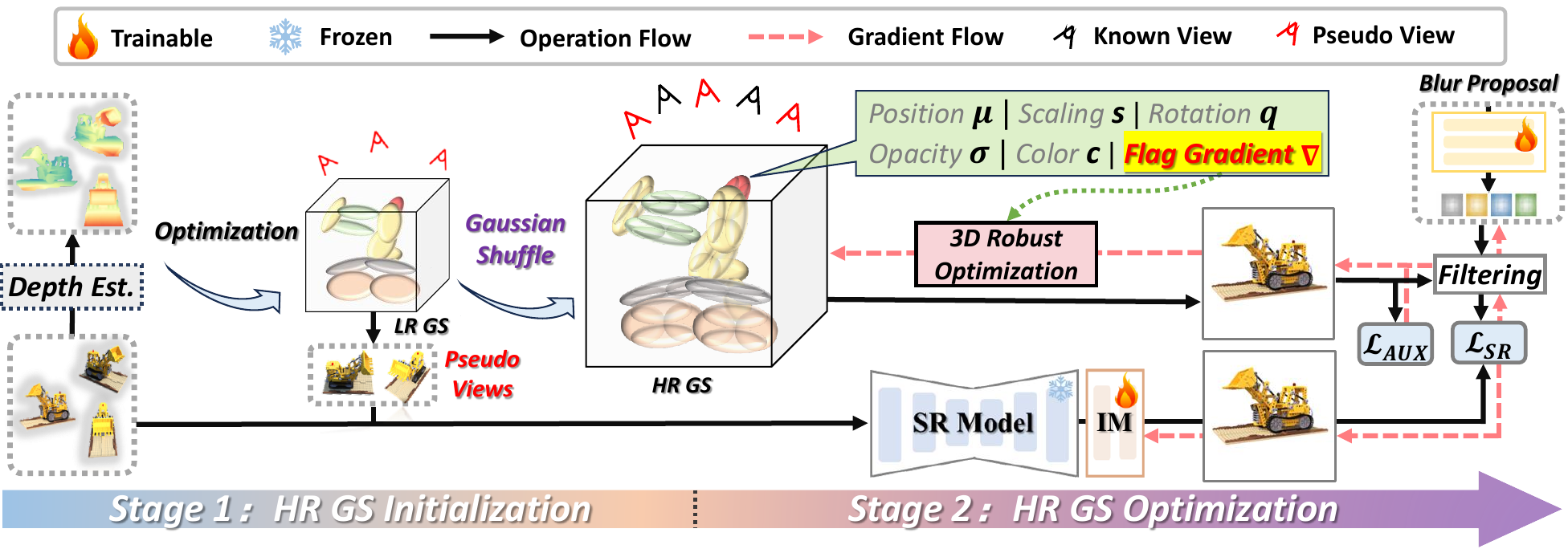}
	\end{center}
	\vspace{-12pt}
	\caption{\textbf{Overview of S2Gaussian}. The S2Gaussian initially optimizes an LR GS and densifies it to initialize the HR GS through a tailored Gaussian Shuffle Split operation. Then, the original sparse views along with the pseudo views rendered by the LR GS are super-resolved together to refine the high-resolution texture with 3D robust optimization. In which an inconsistency modeling module (IM) and a blur proposal module are incorporated to mitigate inconsistency and blurriness, aiming to create 3D scenes with high-fidelity texture details. }
	\label{figure2}
	\vspace{-10pt}
\end{figure*}
\subsection{Super-Resolution Novel View Synthesis}
Super-resolution novel view synthesis aims to reconstruct high-resolution 3D scenes utilizing only low-resolution multi-view inputs. Unlike anti-aliasing methods \cite{barron2021mip,barron2022mip,yu2024mip} which are incapable of rendering intricate details of the novel views and are limited by the input image's information level, this field is more focused on fine-grained and detailed high-resolution rendering of potential high-resolution scenes. As a pioneer in the field, NeRF-SR \cite{wang2022nerf} optimizes high-resolution NeRF 
through a super-sampling strategy, ensuring that the values of low-resolution pixels match the mean value of high-resolution sub-pixels. Follow-up works are designed to utilize either high-resolution reference images \cite{huang2023refsr} or pre-trained 2D models \cite{han2023super,lee2024disr,yoon2023cross} to generate multi-view consistent details. Concurrently, SRGS \cite{feng2024srgs} and GaussianSR \cite{yu2024gaussiansr} are proposed to utilize 2D super-resolution models for texture injection or diffusion prior exploration, respectively. SuperGaussian \cite{shen2024supergaussian} further proves a profile framework to repurpose video upsampling models to achieve 3D super-resolution.

Different from the aforementioned methods, in this work, we ambitiously tackle the challenge of joint sparse-view and super-resolution novel view synthesis by proposing a novel two-stage framework, which cleverly solves the problem of perspectives and clarity deficient to reconstruct geometry-accurate and fine-detailed 3D scenes, making it more desirable and feasible in real-world applications.

\section{Methodology}
\subsection{Preliminaries}
\noindent\textbf{3D Gaussian Splatting} (3DGS)
as an emerged powerful technique for novel view synthesis, models a 3D scene using a collection of $G$ colored Gaussians:
\begin{equation}
	\setlength{\abovedisplayskip}{1pt}
	\setlength{\belowdisplayskip}{1pt}
	\resizebox{.7\hsize}{!}{$
		\begin{aligned} 
		g_i(x) = \exp\left(-\frac{1}{2} (x - \mu_i)^\top \Sigma_i^{-1} (x - \mu_i)\right),
		\end{aligned}$}
\end{equation}
where \(1 \leq i \leq G\), \(\mu_i \in \mathbb{R}^3\) represents the mean or center of the Gaussian, and \(\Sigma_i \in \mathbb{R}^{3 \times 3}\) denotes Gaussian's covariance matrix, specifying its shape and size, which is described by scaling vector $s \in \mathbb{R}^3_+$ and rotation vector $q \in \mathbb{R}^4$ for implementation. Additionally, each Gaussian is characterized by an opacity \(\sigma_i \in \mathbb{R}_+\) and a view-dependent color \(c_i(v) \in \mathbb{R}^{k_c}\) (where $k_c$ is changed) for rasterization.

To faithfully reconstruct the 3D scene, 3DGS also employs densification operation, i.e., the split, clone, and prune operation, to adaptively regulate the number and density of Gaussian primitives. For simplicity, we deploy the same adaptive density control as the original 3DGS and will not elaborate further in the subsequent presentation.
\vspace{-3pt}
\subsection{S2Gaussian Overview}
\vspace{-3pt}

The schematic illustration of the proposed S2Gaussian is depicted in Fig. \ref{figure2}. S2Gaussian is primarily composed of two phases i.e., the HR GS initialization stage and the HR GS optimization stage. In the HR GS initialization stage, we first optimize a low-resolution Gaussian representation using the sparse views as well as the predicted depth information, which is further utilized to initialize the high-resolution Gaussian representation by densifying and detailing the Gaussian primitives through our tailored Gaussian Shuffle Split operation. In the HR GS optimization stage, we further optimize the high-resolution Gaussians with the super-resolved images generated by pre-trained super-resolution model from both original sparse views and the pseudo-views rendered by LR Gaussians. In particular, a dedicated blur-free inconsistency modeling scheme and a 3D robust optimization strategy are advanced to eliminate the effects of multi-view inconsistency and imperfect supervision to reconstruct more detailed and high-quality 3D scenes.

\subsection{Stage 1: HR GS Initialization}
Directly reconstructing a high-resolution 3D scene from sparse low-resolution viewpoints is extremely challenging, as it is not only prone to overfitting specific viewpoints but also lacks high-resolution detail constraints. To alleviate such a reconstruction dilemma, we advocate initially reconstructing a low-resolution Gaussian representation as an auxiliary to facilitate the optimization of high-resolution Gaussians. On the one hand, the low-resolution Gaussian can be used to initialize high-resolution Gaussians to provide reliable base Gaussian ellipsoids and 3D scene structures. On the other hand, it can provide pseudo-views as additional supervision to avoid overfitting on specific viewpoints as well as provide more potential details and textures for high-resolution scenes.

Concretely, we first obtain the monocular depth maps for the sparse low-resolution training views by utilizing the handy pre-trained depth estimation model \cite{ranftl2021vision}. Then we utilize the RGB images as well as the depth maps to co-optimize a low-resolution Gaussian representation with mature depth regularization technologies \cite{li2024dngaussian,zhu2025fsgs}, where Pearson correlation loss \cite{zhu2025fsgs} is employed in this paper as a baseline. Thereafter, a tailored Gaussian Shuffle Split regime is proposed to densify and detail the sparse Gaussian primitives so as to initialize HR Gaussians for better high-resolution fine-grained details representation.

\noindent\textbf{Gaussian Shuffle Split.}
A precise and detailed initialization is crucial for reconstructing scene details \cite{kerbl20233d}. Although the structure and layout of the entire 3D scene are basically guaranteed after the optimization of the low-resolution views, the initialized sparse and coarse Gaussian ellipsoids struggle to simulate high-resolution details which require denser
Gaussians for finer representation \cite{yan2024multi}. Even equipped with adaptive density control \cite{kerbl20233d}, it falls short in our settings as there are neither dense perspectives nor high-quality details for supervision. To address this issue,  we propose a training-free local Gaussian densification strategy \emph{Gaussian Shuffle Split} which can provide more mobilizable Gaussian primitives by replacing the original large Gaussian with six small Gaussians to facilitate more comprehensive simulation of fine-grained details and textures in high-resolution scenes. 

\begin{figure}[t]
	\begin{center}
		\includegraphics[width=\linewidth]{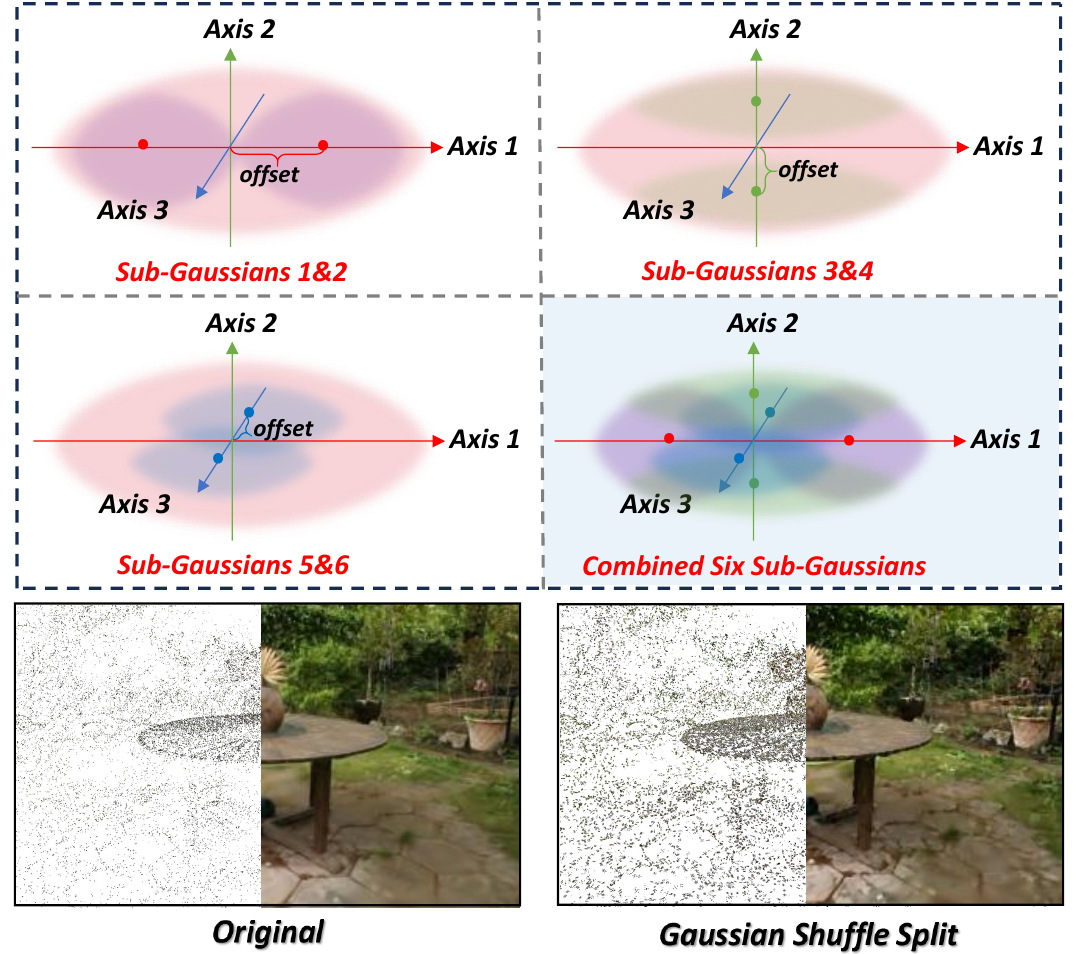}
	\end{center}
	\vspace{-10pt}
	\caption{Illustration of Gaussian Shuffle Split that utilizes a combination of six compact Gaussians to replace the original large Gaussian primitive, which can densify the 3D scene virtually without damaging the original 3D representation. }
	\label{figure3}
	\vspace{-17pt}
\end{figure}

As illustrated in Fig. \ref{figure3}, given a Gaussian primitive with attributes \(\{\mu, s, q, \sigma, c\}\), we first generate six duplicates, each center shifted along one of the six directions aligned with the original Gaussian's three principal axes, i.e., the positive and negative directions of each axis. The shift for each axis is set to \(\alpha\) (0.5 by default) times the corresponding axis's scaling $s=[s_1, s_2, s_3]$ value. Thus the new position of the six sub-Gaussians can be obtained: 
\begin{equation}
	\setlength{\abovedisplayskip}{2pt}
	\setlength{\belowdisplayskip}{1pt}
	\resizebox{\linewidth}{!}{$
	\begin{aligned} 
	&\mu_{sub1} \!=\! \mu \!+\! R \otimes [\alpha s_1,0,0]^T ,
	\mu_{sub2} \!=\! \mu \!+\! R \otimes [-\alpha s_1,0,0]^T, \\
	&\mu_{sub3} \!=\! \mu \!+\! R \otimes [0,\alpha s_2,0]^T, 
	\mu_{sub4} \!=\! \mu \!+\! R \otimes [0,-\alpha s_2,0]^T, \\
	&\mu_{sub5} \!=\! \mu \!+\! R \otimes [0,0,\alpha s_3]^T, 
	\mu_{sub6} \!=\! \mu \!+\! R \otimes [0,0,-\alpha s_3]^T, \\
	\end{aligned}$}
\end{equation}
 where $R$ represents the rotation matrix calculated from $q$ to align with the original rotation of the Gaussian.
Additionally, for each shifted sub-Gaussian, we scale their \( s \) value corresponding to the offset axis to \( \frac{1}{4} \) of its original value, i.e., reducing the scale along that axis to shrink the size. The remaining two \( s \) values are scaled by an another factor of \( \lambda \) relative to their original values:
\begin{equation}
	\setlength{\abovedisplayskip}{2pt}
	\setlength{\belowdisplayskip}{2pt}
	\begin{aligned} 
	s_{sub1} = s_{sub2}  = [s_1/4, s_2/\lambda, s_3/\lambda], \\
		s_{sub3} = s_{sub4}  = [s_1/\lambda, s_2/4, s_3/\lambda], \\
			s_{sub5} = s_{sub6}  =[s_1/\lambda, s_2/\lambda, s_3/4],
	\end{aligned}
\end{equation}
 where $\lambda
$ is experimentally set to 1.9 which we found this ratio ensures that the combined representation of the sub-Gaussians better approximate the original Gaussian, preserving the integrity of the 3D scene.
All other attributes, i.e., rotation $q$, opacity $\alpha$, and color $c$ remain identical to the original Gaussian ellipsoid. The combined six sub-Gaussians are utilized to substitute the original large Gaussians for densified 3D representation.

 We deployed Gaussian Shuffle Split only for Gaussian primitives with opacity greater than 0.5 since these are more likely to be responsible for representing object surfaces and key structures which require a more detailed representation in high-resolution scenes.
The opacity of all primitives is set close to zero after the Gaussian Shuffle Split to facilitate the automatic elimination of redundant Gaussians with low opacity.

\subsection{Stage 2: HR GS Optimization}
With the initialized high-resolution Gaussians, in this stage, we aim to continue optimizing it with the super-resolved images generated from both original sparse views and the pseudo-views rendered by the LR Gaussians. We synthesize pseudo views by interpolating virtual views between two known views as in \cite{zhu2025fsgs}. Considering the prohibitively expensive data costs associated with 3D super-resolution models and their computational complexity, we advocate for utilizing off-the-shelf pre-trained 2D super-resolution models to enhance low-resolution views.
Unfortunately, directly leveraging 2D super-resolution model cannot ensure multi-view consistency, and the low-resolution pseudo-views inevitably exhibit artifacts since there are areas that have not been optimized before. The Gaussian primitives, in their attempt to accommodate these inconsistencies and erroneous representations, will lead to the optimized 3D scene converging toward inaccurate structures and blurriness. 

During optimization, two crucial aspects contribute fundamentally to these problems, i.e., inconsistent supervision and erroneous updates caused by artifacts. Accordingly, a customized blur-free inconsistency modeling scheme and a 3D robust optimization strategy are proposed in the
following to alleviate potential inaccurate reconstruction.

\noindent\textbf{Blur-Free Inconsistency Modeling.}
To mitigate the inconsistency effort of the single-image super-resolution model, we employ a learnable inconsistency modeling module (IM), i.e., two residual blocks \cite{he2016deep}, after the pre-trained super-resolution model to simulate the inconsistency across different views $I_{SR}^{IM} = I_{SR} + IM(I_{SR})$, thus avoiding making Gaussians to represent such inconsistency. 
However, we empirically found that this module tends to lose details during optimization to win better consistency, resulting in the loosing of sharp textures.
Therefore, we further propose a blur proposal module (BP) to blur the rendered image which is then constrained with modified super-resolution image $I_{SR}^{IM}$,
thereby compensating for the loss of fine-grained details and textures caused by IM. The blur proposal module, i.e., a four-layer convolutional network, predicts per-pixel blur kernels $\mathcal{B}_k \in \mathbb{R}^{K\times K \times H \times W}$ ($K$ is the blur kernel size, 5 for $4\times$ task) with the   gradient detached rendered image $R_{HR}^{detach}$ as input $\mathcal{B}_k = BP(R_{HR}^{detach})$. Then the blurred high-resolution image is obtained via $R_{HR}^{blur} = R_{HR} * \mathcal{B}_k$ which is utilized to compute loss with $I_{SR}^{IM}$, avoiding the smoothing imposed by directly constraining $R_{HR}$:
\begin{equation}
	\setlength{\abovedisplayskip}{1pt}
	\setlength{\belowdisplayskip}{1pt}
	\begin{aligned} 
		\mathcal{L}_{SR} \!=\! (1\!-\! \beta) \mathcal{L}_{1}(R_{HR}^{blur},I_{SR}^{IM}) \!+\! \beta \mathcal{L}_{D\!-\!SSIM}(R_{HR}^{blur},I_{SR}^{IM}) ,
	\end{aligned}
\end{equation} 
where $\beta$ is set to 0.2 similar to the original 3DGS \cite{kerbl20233d}.

\begin{figure}[t]
	\begin{center}
		\includegraphics[width=\linewidth]{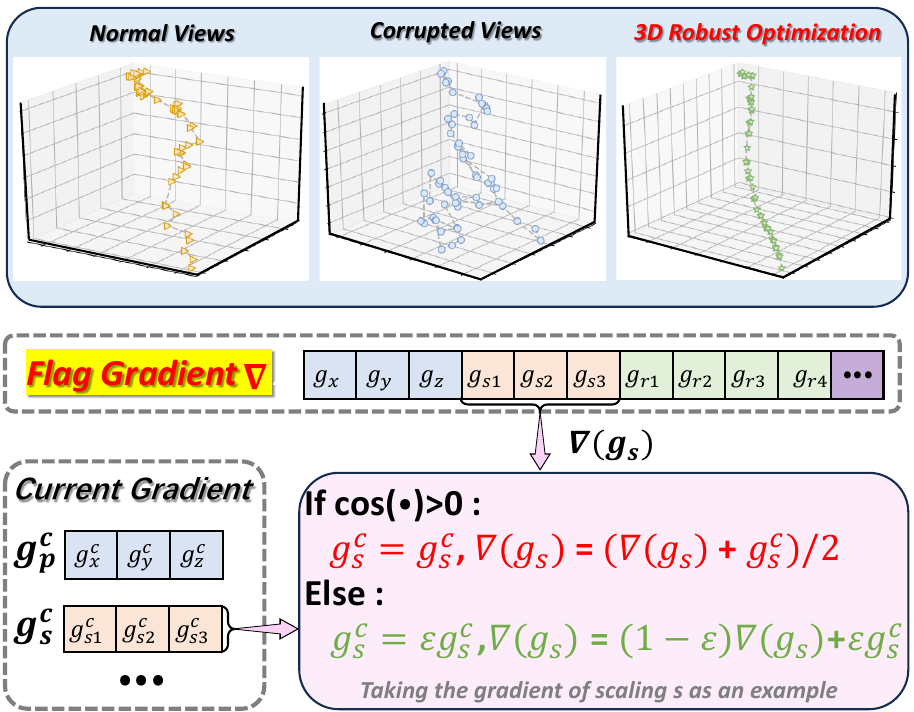}
	\end{center}
	\vspace{-10pt}
	\caption{\emph{Top:} Visualization of the gradient variation under different quality views supervision and our proposed 3D robust optimization. \emph{Bottom:} Illustration of the proposed 3D robust optimization strategy that takes the gradient of scaling $s$ as an example.}
	\label{figure4}
	\vspace{-11pt}
\end{figure}

\noindent\textbf{3D Robust Optimization.} Albeit the above scheme could surmount the detrimental effects of multi-view inconsistencies, the optimization of Gaussians still unavoidably suffers from improper supervision of pseudo-views where some areas are not sufficiently optimized.
To this end, we empirically observed that optimizing Gaussian primitives utilizing normal high-quality views results in stable and overall consistent gradients (top left of Fig. \ref{figure4}) whereas corrupted views can lead to significant perturbation and gradient confusion (top middle of Fig. \ref{figure4}), ultimately resulting in blurry rendering. Intrigued by
this observation, we aim to design a more robust optimization strategy to mitigate the fluctuations caused by imperfect supervision.

\begin{table*}[ht!]
	\centering
	\caption{Quantitative comparison on Blender $\times$4 ~(8 views), LLFF $\times$4 ~(3 views), and Mip-NeRF360 $\times$4 ~(24 views). The best, second best, and third best entries are marked in red, orange, and yellow, respectively. }
	\setlength{\tabcolsep}{2mm}
	\footnotesize
	\begin{tabular}{l|cccc|cccc|cccc}
		\toprule[1pt]
		\multirow{2}{*}{\textbf{Method}} & \multicolumn{4}{c|}{Blender $\times$4 ~(8 views)} & \multicolumn{4}{c|}{ LLFF $\times$4 ~(3 views)} & \multicolumn{4}{c}{Mip-NeRF 360 $\times$4 ~(24 views)}  \\
		%\cmidrule(lr){2-} 
		& PSNR$\uparrow$ & SSIM$\uparrow$ & LPIPS$\downarrow$& FID$\downarrow$ & PSNR$\uparrow$ & SSIM$\uparrow$ & LPIPS$\downarrow$& FID$\downarrow$ & PSNR$\uparrow$ & SSIM$\uparrow$ & LPIPS$\downarrow$ & FID$\downarrow$ \\
		\midrule
		NeRF-SR \cite{wang2022nerf}& 12.41&0.744&0.515&93.79&9.28&0.226&0.617&169.58&10.26&0.269&0.628&151.62 \\
		RegNeRF \cite{niemeyer2022regnerf}& 20.68&0.841&0.129&41.28&15.78&0.432&0.422&116.74&17.28&0.417&0.449&105.45 \\
		SparseNeRF \cite{wang2023sparsenerf}&21.32 &0.844&0.136&42.62&16.32&0.446&0.428&127.64&17.38&0.432&0.446& 109.27\\
		\midrule
		
		3DGS \cite{kerbl20233d}&20.65&0.831&0.141&42.58&13.01&0.272&0.489&138.53&16.25&0.342&0.474&108.32 \\
		SRGS \cite{feng2024srgs} &22.64&0.862&\cellcolor{yellow! 30}0.116&36.24&17.58&0.472&0.291&102.43&18.37&0.457&\cellcolor{yellow! 30}0.397&91.62\\
		Mip-Splatting \cite{yu2024mip} &\cellcolor{orange!30}22.78 &\cellcolor{orange!30}0.868&0.119&\cellcolor{orange!30}33.63&15.65&0.433&0.372&114.03&18.08&0.448&0.421&\cellcolor{yellow! 30}86.17\\
		DNGaussian \cite{li2024dngaussian}&20.97 &0.844&0.135&41.15&16.27&0.430&0.346&129.26&17.21&0.422&0.452&112.22 \\
		FSGS \cite{zhu2025fsgs} & 21.45 &0.854&0.133&40.36&\cellcolor{yellow! 30}18.35&\cellcolor{yellow! 30}0.505&\cellcolor{yellow! 30}0.247&\cellcolor{yellow! 30}97.07&\cellcolor{yellow! 30}18.64&\cellcolor{yellow! 30}0.458&\cellcolor{orange!30}0.394&94.21\\
		FSGS \cite{zhu2025fsgs}+SRGS \cite{feng2024srgs} &\cellcolor{yellow! 30}22.68 &\cellcolor{yellow! 30}0.865&\cellcolor{orange!30}0.114&\cellcolor{yellow! 30}35.42&\cellcolor{orange!30}19.49&\cellcolor{orange!30}0.590&\cellcolor{orange!30}0.175&\cellcolor{orange!30}62.63&\cellcolor{orange!30}19.83&\cellcolor{orange!30}0.513&0.448&\cellcolor{orange!30}84.12 \\
		\textbf{S2Gaussian (Ours)} &  \cellcolor{red!30}\textbf{24.19}&\cellcolor{red!30}\textbf{0.879}&\cellcolor{red!30}\textbf{0.089}&\cellcolor{red!30}\textbf{28.47}&\cellcolor{red!30}\textbf{20.45}&\cellcolor{red!30}\textbf{0.654}&\cellcolor{red!30}\textbf{0.139}&\cellcolor{red!30}\textbf{45.89}&\cellcolor{red!30}\textbf{22.05}&\cellcolor{red!30}\textbf{0.687} & \cellcolor{red!30}\textbf{0.296}&\cellcolor{red!30}\textbf{43.51} \\
		\bottomrule[1pt]
	\end{tabular}
	\label{table1}
	\vspace{-7pt}
\end{table*}

More specifically, we augment the canonical
Gaussian primitives
\begin{equation}
	\setlength{\abovedisplayskip}{1pt}
	\setlength{\belowdisplayskip}{1pt}
	\begin{aligned} 
	\{\mu, s, q, \sigma, c\} \cup \{ \nabla \},
	\end{aligned}
\end{equation}
with an additional per-Gaussian attribute flag gradient $\nabla \in \mathbb{R}^{11+k_c}$. Where $\nabla$ is used to record the gradient trends of other attributes of the Gaussian, i.e., $\mu, s, q, \sigma,$ and the changed $k_c$ SH coefficients. 
As shown in Fig. \ref{figure4}, for each Gaussian in one optimization step, we first compute the cosine similarity between the current gradient for each attribute $g_i^c$ ($i$ indicate different attribute) and the corresponding flag gradient $\nabla(g_i)$ stored in $\nabla$.
If the cosine similarity is greater than zero, indicating that the gradient update trends are aligned, the gradient $g_i^c$ will be directly used to optimize the parameter, and the flag gradient value $\nabla(g_i)$ will be updated to the average one:
\begin{equation}
	\setlength{\abovedisplayskip}{1pt}
	\setlength{\belowdisplayskip}{1pt}
	\begin{aligned} 
		g_i^c = g_i^c , ~\nabla(g_i) = (\nabla(g_i) +g_i^c) /2 .
	\end{aligned}
\end{equation}
Conversely, if the cosine similarity is less than or equal to zero, indicating potential disturbances, the current gradient will be scaled down by a factor of $\epsilon$ to slow the parameter updates, and $\nabla(g_i)$ will be updated as follows:
\begin{equation}
	\setlength{\abovedisplayskip}{1pt}
	\setlength{\belowdisplayskip}{1pt}
	\begin{aligned} 
		g_i^c = \epsilon g_i^c , ~\nabla(g_i) = (1-\epsilon) \nabla(g_i) + \epsilon g_i^c,
	\end{aligned}
\end{equation}
where $\epsilon$ is set to 0.1 to attenuate potentially erroneous updates and preserve the ability to shift update trends. In practice, such strategy can be combined with any gradient-based optimizer, e.g., Adam, by simply passing the modified gradients to the respective optimizer. 
Notably, $\nabla$ will be eliminated after the optimization and will not affect the rendering speed and storage consumption.

As illustrated in the top right of Fig. \ref{figure4}, our proposed 3D robust optimization scheme can significantly improve the stability of Gaussian primitive optimization by eliminating erroneous updates and perturbations thus reconstructing more accurate and high-quality 3D representations.

\noindent\textbf{Total Objective.}
Apart from $\mathcal{L}_{SR}$, an auxiliary loss combined with total variation (TV) loss and sub-pixel constraint is also deployed to stabilize the training:
\begin{equation}
	\setlength{\abovedisplayskip}{1pt}
	\setlength{\belowdisplayskip}{1pt}
	\begin{aligned} 
	\mathcal{L}_{AUX} = \mathcal{L}_{TV}(R_{HR}) + \mathcal{L}_{1}(R_{HR} \downarrow ,I_{LR}),
	\end{aligned}
\end{equation}
where $\downarrow$ indicates area average downsampling and $I_{LR}$ denotes the low-resolution images before super-resolution.

 In summary, the final loss $\mathcal{L}$ for high-resolution Gaussians optimization is defined as: 
\begin{equation}
	\setlength{\abovedisplayskip}{1pt}
	\setlength{\belowdisplayskip}{1pt}
	\begin{aligned} 
		\mathcal{L} = \mathcal{L}_{AUX} + \mathcal{L}_{SR}.
	\end{aligned}
\end{equation}

\vspace{-8pt}
\section{Experiments}
\vspace{-3pt}
Following prior literature \cite{lee2024disr,yu2024gaussiansr}, we primarily experiment on 4 $\times$ super-resolution tasks in the maintext since most real-world 2D super-resolution models \cite{yue2024resshift,wang2024exploiting} are mainly trained on $\times$ 4 data, more diagnostic experiments on other scaling factors are provided in the Suppl.

\begin{figure*}[h]
	
	\begin{center}
		\includegraphics[width=\linewidth]{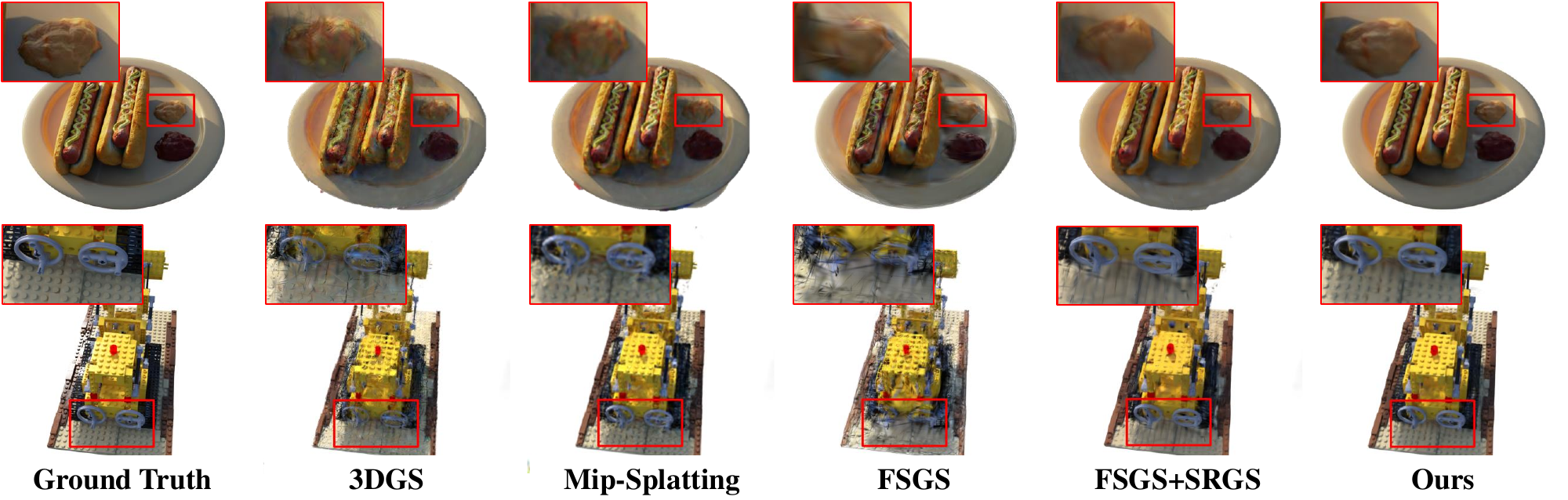}
	\end{center}
	\vspace{-15pt}
	\caption{Qualitative comparisons on Blender $\times$4 datasets with 8 input views. Our method produces more visually appealing results, successfully reconstructing intricate thin structures with fine-grained details. \textbf{Best viewed at screen!} }
	\label{figure6}
		\vspace{-4pt}
\end{figure*}
\begin{figure*}[h]
	
	\begin{center}
		\includegraphics[width=\linewidth]{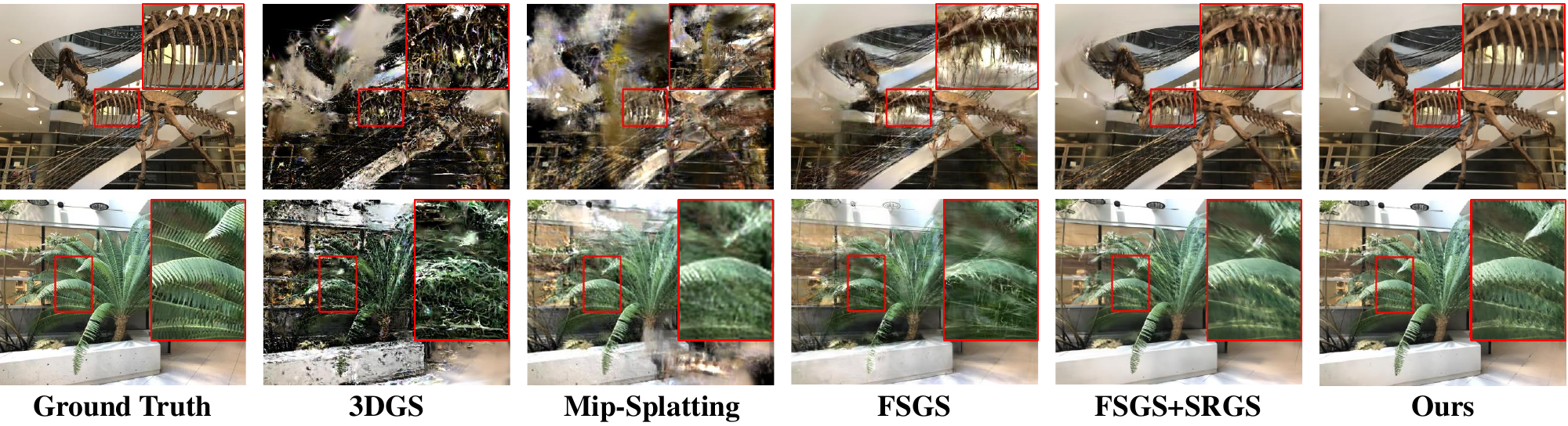}
	\end{center}
	\vspace{-15pt}
	\caption{Qualitative comparisons on LLFF $\times$4 datasets with 24 input views. Our method still generates the most accurate super-resolution 3D scenes, preserving both photo-realistic texture and geometric-level details. \textbf{Best viewed at screen!} }
	\label{figure7}
		\vspace{-10pt}
\end{figure*}
\begin{figure*}[h]
	
	\begin{center}
		\includegraphics[width=\linewidth]{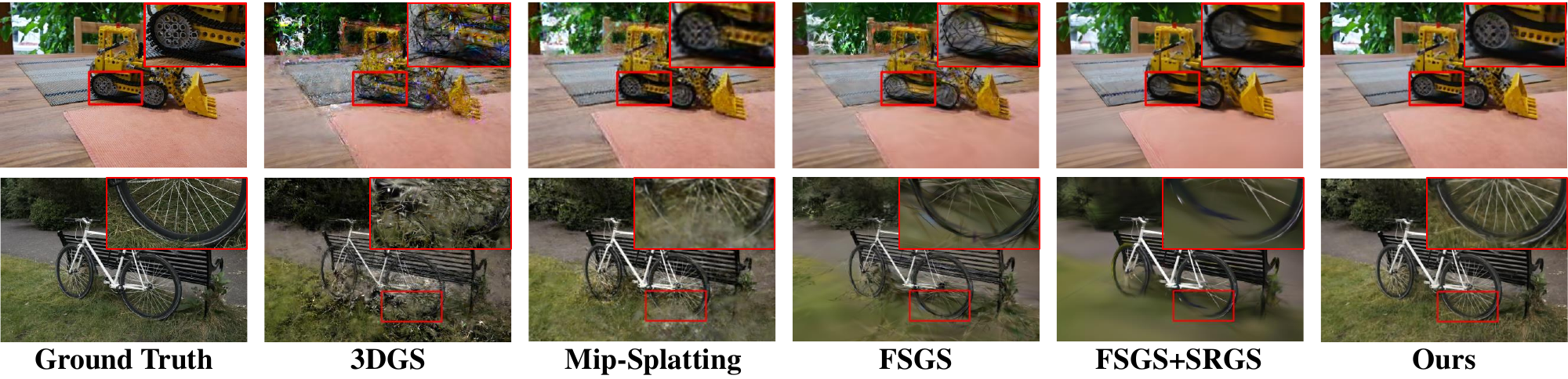}
	\end{center}
	\vspace{-15pt}
	\caption{Qualitative comparisons on Mip-NeRF360 $\times$4 datasets with 24 input views. Our method consistently delivers more coherent fine structures than other methods in
		large-scale scenes. \textbf{Best viewed at screen!} }
	\label{figure8}
		\vspace{-10pt}
\end{figure*}

\subsection{Implementation Details}
We implemented S2Gaussian upon the original 3DGS repository and all experiments are conducted on a single RTX3090 GPU. For the optimization of LR GS, the pre-trained DPT \cite{ranftl2021vision} model is adapted for depth estimation and the total iterations are set to 10,000. During the HR GS optimization stage, the Gaussian primitives are further optimized with additional 10,000 iterations. As for the off-the-shelf 2D super-resolution model, we opt for powerful diffusion-based ResShift \cite{yue2024resshift} as our backbone. 
We conduct experiments on three benchmark datasets with SfM for initialization, advanced techniques like DUSt3R \cite{wang2024dust3r} can facilitate the initialization but that is not the main purpose of our current work.

\noindent\textbf{Blender Datasets} \cite{mildenhall2021nerf} contain 8 detailed objects synthesized by Blender with a resolution of 800 $\times$ 800. 
Each object initially includes 100 training images and 200 testing images. We uniformly sample 8 images from the original 100 training images and use the low-resolution of $200\times200$ pixels for training.

\noindent\textbf{LLFF Datasets} \cite{mildenhall2019local} consist of 8 forward-facing real-world scenes. Following the previous setup, we select every eighth image as the test set, and evenly sample 3 sparse views from the remaining images for training. We leverage low-resolution images at $252 \times 189$ pixels for training and render at a higher resolution of $1008 \times 756$ pixels.

\noindent\textbf{Mip-NeRF 360 Datasets} \cite{barron2022mip} consist of 9 real-world scenes with 5 outdoors and 4 indoors. Similar to LLFF, we use 1/8 of the full view for testing and uniformly sample 24 images from the remaining views for training. We utilize the images with downsampling rate of 4 (about 1K resolution) as the target size and further downsample the training images by a factor 4 to obtain sparse low-resolution inputs. 

The following measures are used to evaluate the quantitative performance: Peak Signal-to-Noise Ratio (PSNR) \cite{huynh2008scope}, Structural Similarity Index (SSIM) \cite{wang2004image}, LPIPS (VGG) \cite{zhang2018unreasonable}, and Fréchet Inception Distance (FID) \cite{heusel2017gans}.

\subsection{Quantitative and Qualitative Comparisons}
To rigorously validate the effectiveness of our proposed S2Gaussian, we conduct a comparative analysis against a range of prior methods, including NeRF-SR \cite{wang2022nerf}, RegNeRF \cite{niemeyer2022regnerf}, SparseNeRF \cite{wang2023sparsenerf}, vanilla 3DGS \cite{kerbl20233d}, SRGS \cite{feng2024srgs}, Mip-Splatting \cite{yu2024mip}, DNGaussian \cite{li2024dngaussian}, and FSGS \cite{zhu2025fsgs}. Besides, we also include an enhanced version of FSGS with SRGS to provide a
standard upsampling baseline. While NeRF-SR and FSGS+SRGS employ super-resolution techniques directly, the remaining methods are trained on low-resolution input views and subsequently rendered at higher resolutions.

\noindent\textbf{Quantitative Evaluation.} Tab. \ref{table1} presents quantitative comparison results for 4 $\times$ sparse-view super-resolution novel view synthesis task on the Blender \cite{mildenhall2021nerf}, LLFF \cite{mildenhall2019local}, and Mip-NeRF 360 datasets \cite{barron2022mip}. It is observed that our S2Gaussian delivers remarkable performance gains and outperforms all competitive method significantly in terms of PSNR, SSIM, LPIPS, and FID metrics. Especially for Mip-NeRF 360 datasets, S2Gaussian exceeds the top-performing combinatorial method FSGS+SRGS by an impressive margin of 2.22 dB PSNR, underscoring its superior capability in reconstructing large, real-world scenes. In addition to delivering higher reconstruction fidelity, reflected by PSNR and SSIM, S2Gaussian also achieves substantial improvements in perceptual metrics LPIPS and FID, validating the effectiveness of S2Gaussian in converging towards greater perception quality with faithful and fidelity details.

\noindent\textbf{Qualitative Evaluation.}
We also demonstrate visual comparisons in Fig. \ref{figure6}, Fig. \ref{figure7}, and Fig. \ref{figure8}. As suggested, both 3DGS and Mip-Splatting struggle with the extremely sparse view problem, while FSGS is limited to characterizing high-resolution details. Despite the combinatorial method FSGS+SRGS can reconstruct a reasonable overall structure but suffers from inconsistency and sparsity leading to missing details and blurring. In comparison, S2Gaussian recovers more coherent fine structures with natural textures and harmonious global tone while other methods tend to introduce improper structures and unnatural details, especially for the extremely sparse scenes (Fig. \ref{figure7}) as well as large-scale scenarios (Fig. \ref{figure8}).

\subsection{Ablation Studies}

In Tab. \ref{table_pc}, we conduct ablation experiments on the components introduced in S2Gaussian. The effectiveness of each proposed component in S2Gaussian is evaluated by gradually integrating them into the model, revealing their individual contributions to the overall performance. More detailed analyses are as below.

\vspace{-7pt}
\begin{table}[h]	
	\footnotesize
	\caption{Ablation studies on Mip-NeRF 360 $\times$4 ~(24 views).}
	\tabcolsep=2.7pt
	\vspace{-19pt}
	\begin{center}
		\begin{tabular}{lcccc}
			\toprule[1pt]
			$~~~$ Variant&PSNR$\uparrow$&SSIM$\uparrow$&LPIPS$\downarrow$&FID$\downarrow$\\
			\midrule
			$~~~$ Baseline &19.79&0.520&0.429&87.22\\
			$+$ Two-Stage 	&20.06&0.528&0.415&80.64\\
			$+$ Gaussian Shuffle Split	&20.62&0.559&0.376&71.29\\
			$+$	Blur-Free Inconsistency Modeling &21.12&0.624&0.346&59.63\\
			$+$ 3D Robust Optimization&22.05&0.687&0.296&43.51\\
			\bottomrule[1pt]
		\end{tabular}
	\end{center}
	\label{table_pc}
	\vspace{-15pt}
\end{table} 

\noindent\textbf{Effect of Gaussian Shuffle Split.} As shown in Tab. \ref{table_1}, fewer sub-Gaussians lead to suboptimal detail modeling (lower PSNR), while 6 offers the best trade-off, while more sub-Gaussians will not only increase the optimization complexity but also introduce extra memory overhead. Additionally, 6 sub-Gaussians can be symmetrically located along the three principal axes of the original Gaussian, perfectly approximating the original Gaussian representation.
\begin{table}[h]
	\vspace{-12pt}
	\centering
	\tabcolsep=2.6pt
	\makeatletter\def\@captype{table}\makeatother
	\renewcommand\arraystretch{0.95}
	\caption{Ablation of sub-Gaussian number on Mip-NeRF 360.}
	\vspace{-10pt}
	\small
	\begin{tabular}{ccccccc}
			\toprule[1pt]
		Number&4&5&\textbf{6}&7&8&9\\
		\hline
		Storage Memory (M)&67&89&\textbf{112}&137&169&202\\
		Final PSNR&21.76&21.95&\textbf{22.05}&22.03&22.09&22.08\\
	\bottomrule[1pt]
	\end{tabular}
	\label{table_1}
	%\end{minipage}
	\vspace{-12pt}
\end{table}

\begin{figure}[h]
	\begin{center}
		\includegraphics[width=\linewidth]{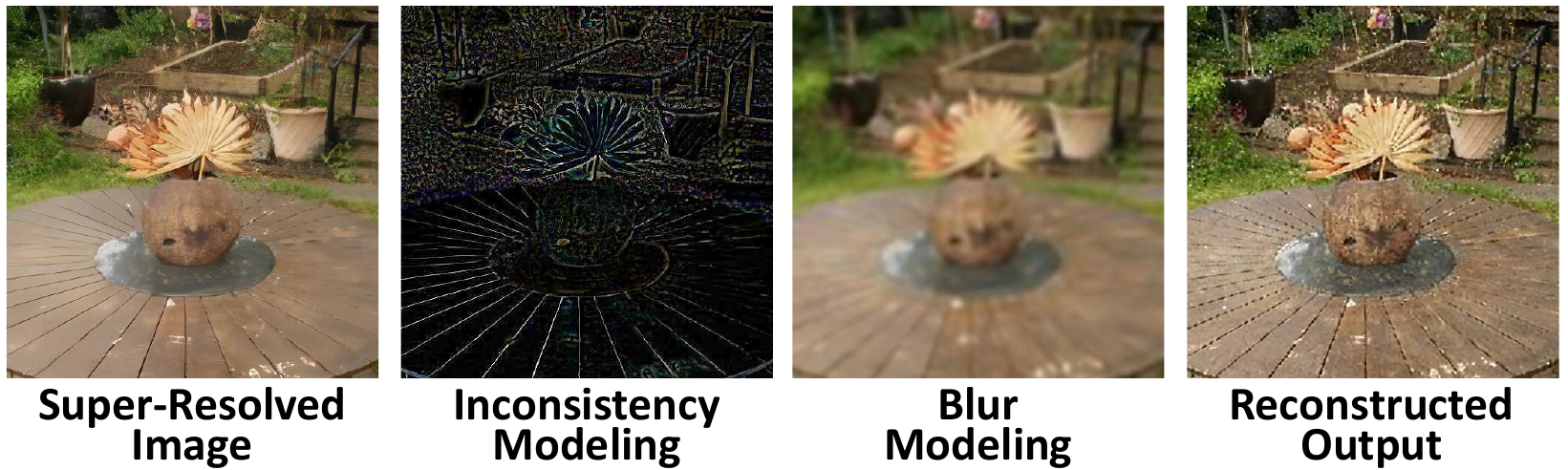}
	\end{center}
	\vspace{-15pt}
	\caption{ Visualization of the super-resolved image, the residual output of IM, the blurred reconstructed image with predicted blur kernels, and the original reconstructed output. }
	\label{figure10}
	\vspace{-12pt}
\end{figure}

\noindent\textbf{Effect of Blur-Free Inconsistency Modeling.}
 As illustrated in Fig. \ref{figure10}, IM models the inconsistency of super-resolved images, and while this would introduce blurring, we model this effect explicitly through blur kernel learning, thus reconstructing results that are blur-free and with fine structural details.

\newpage

\begin{figure}[h]
	
	\begin{center}
		\includegraphics[width=\linewidth]{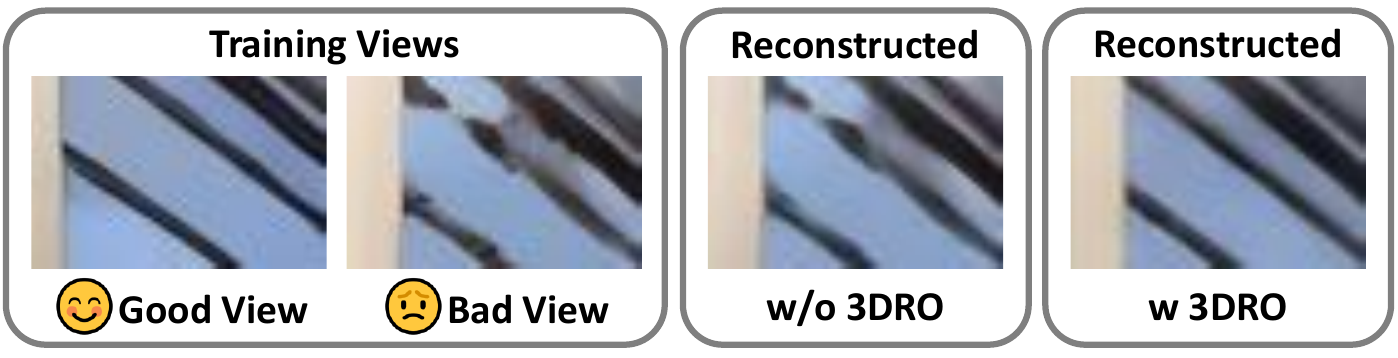}
	\end{center}
	\vspace{-15pt}
	\caption{ Visual comparison with or without 3D robust optimization (3DRO) under corrupted training views. }
	\label{figure11}
	\vspace{-7pt}
\end{figure}

\noindent\textbf{Effect of 3D Robust Optimization.} Fig. \ref{figure11} showcases visual comparison with or without 3D robust optimization under corrupted training views.
It is observed that 3D robust optimization can effectively avoid unfavorable or erroneous supervision to optimize towards higher quality whereas its absence results in Gaussian primitives simulating corrupted representations. 
%------------------------------------------------------------------------
\vspace{-3pt}
\section{Concluding Remarks}
In this paper, we propose S2Gaussian, an innovative framework that can efficiently reconstruct high-quality 3D scenes with only sparse and low-resolution input views. In contrast to existing frameworks, S2Gaussian can handle both view sparse and clarity deficient simultaneously, which is more functional and practical in real-world scenarios. The proposed framework begins by optimizing a low-resolution Gaussian representation, which is then densified to initialize high-resolution Gaussians via tailored Gaussian Shuffle Split operation. Subsequently, it cleverly solves the view sparsity problem by rendering pseudo-views from low-resolution Gaussians for super-resolution supervision. Wherein a customized blur-free inconsistency modeling scheme and a 3D robust optimization strategy are further proposed to mitigate multi-view inconsistency and eliminate erroneous updates for better geometric structure and fine-grained details reconstruction.
Extensive experiments on multiple benchmarks manifest the effectiveness, superiority, and robustness of our method.
We expect this work to provide insights into more complicated 3D reconstruction and steer future research on this Gordian knot. 

\newpage
{
    \small
    \bibliographystyle{ieeenat_fullname}
    \bibliography{cvpr}
}


\end{document}